\documentclass[10pt,twocolumn,letterpaper]{article}

\usepackage{cvpr}              % To produce the CAMERA-READY version
\usepackage{graphicx}
\usepackage{float}
\usepackage{amsmath}
\usepackage{amssymb}
\usepackage{booktabs}
\usepackage{diagbox}
\usepackage{comment}
\usepackage{multirow}
\usepackage{makecell}
\usepackage[dvipsnames]{xcolor}

\usepackage[accsupp]{axessibility} % Improves PDF readability for those with disabilities.

\usepackage[resetlabels,labeled]{multibib}
\newcites{Sup}{Supplementary Reference}

\definecolor{detcolor}{RGB}{183, 148, 145}%{229, 185, 181}
\definecolor{desccolor}{RGB}{147,169,102}
\definecolor{thencolor}{RGB}{0, 0, 255}

\usepackage[pagebackref,breaklinks,colorlinks]{hyperref}

\usepackage[capitalize]{cleveref}
\crefname{section}{Sec.}{Secs.}
\Crefname{section}{Section}{Sections}
\Crefname{table}{Table}{Tables}
\crefname{table}{Tab.}{Tabs.}

\def\cvprPaperID{3282} % *** Enter the CVPR Paper ID here
\def\confName{CVPR}
\def\confYear{2022}

\begin{document}

\title{Decoupling Makes Weakly Supervised Local Feature Better}

\author{Kunhong~Li$^{1,2}$\quad
        Longguang~Wang$^{3}$\quad
        Li~Liu$^{3,4}$\quad 
        Qing~Ran$^5$\quad
        Kai~Xu$^3$\quad
        Yulan~Guo$^{1,2,3}$\thanks{Corresponding author: Yulan Guo (guoyulan@sysu.edu.cn).}\\
        $^1$Sun Yat-Sen University
        \quad
        $^2$The Shenzhen Campus of Sun Yat-Sen University\\
        $^3$National University of Defense Technology
        \quad
        $^4$University of Oulu
        \quad
        $^5$Alibaba Group
        % <-this % stops a space
}
\maketitle

%%%%%%%%% ABSTRACT
\begin{abstract}
Weakly supervised learning can help local feature methods to overcome the obstacle of acquiring a large-scale dataset with densely labeled correspondences. However, since weak supervision cannot distinguish the losses caused by the detection and description steps, directly conducting weakly supervised learning within a \textit{\textbf{joint} training describe-then-detect} pipeline suffers limited performance. In this paper, we propose a \textit{\textbf{decoupled} training describe-then-detect} pipeline tailored for weakly supervised local feature learning. Within our pipeline, the detection step is decoupled from the description step and postponed until discriminative and robust descriptors are learned. In addition, we introduce a line-to-window search strategy to explicitly use the camera pose information for better descriptor learning. Extensive experiments show that our method, namely PoSFeat (Camera \textbf{Po}se \textbf{S}upervised \textbf{Feat}ure), outperforms previous fully and weakly supervised methods and achieves state-of-the-art performance on a wide range of downstream task. \let\thefootnote\relax\footnotetext{Codes: \scriptsize{\href{https://github.com/The-Learning-And-Vision-Atelier-LAVA/PoSFeat}{https://github.com/The-Learning-And-Vision-Atelier-LAVA/PoSFeat}}.}
\end{abstract}

%%%%%%%%% BODY TEXT
\section{Introduction}

\begin{figure}[t]
    \centering
    \subfloat[DISK-W, a DISK model \cite{tyszkiewiczDISKLearningLocal2020a} trained with weak supervision]{
    \label{fig_ambiguity_disk}
	\includegraphics[width=0.45\textwidth]{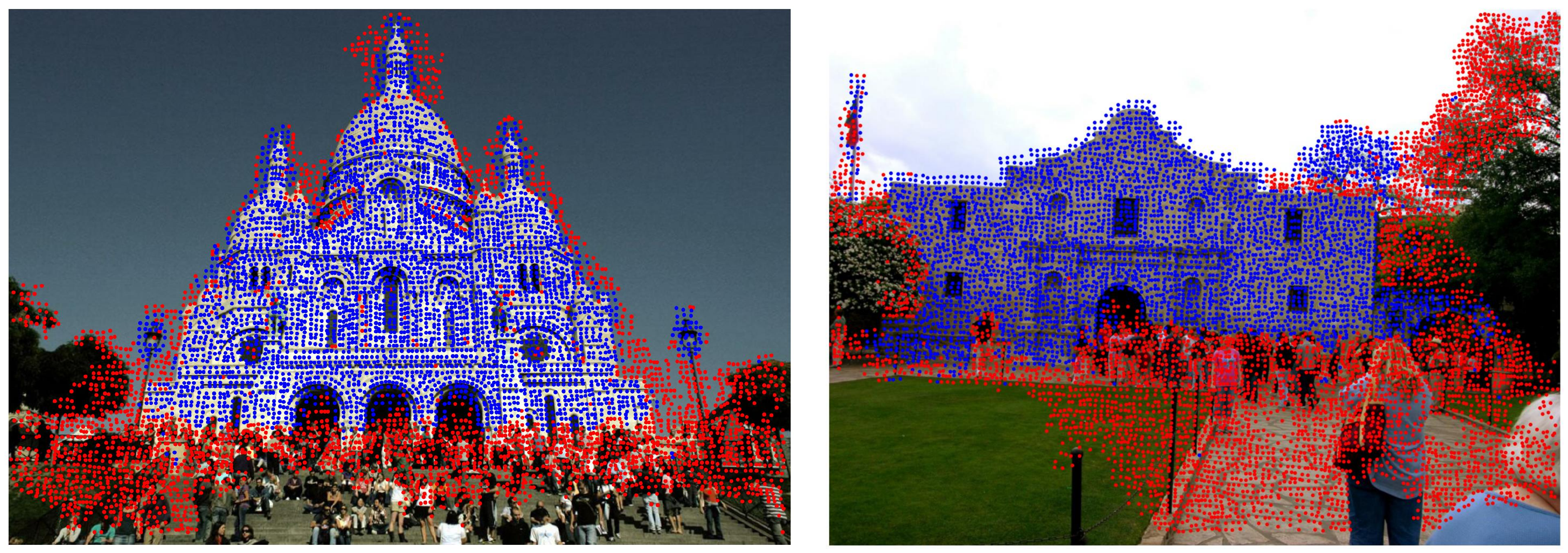}
	} 
    \quad
    \subfloat[PoSFeat (ours)]{
    \label{fig_ambiguity_pos}
	\includegraphics[width=0.45\textwidth]{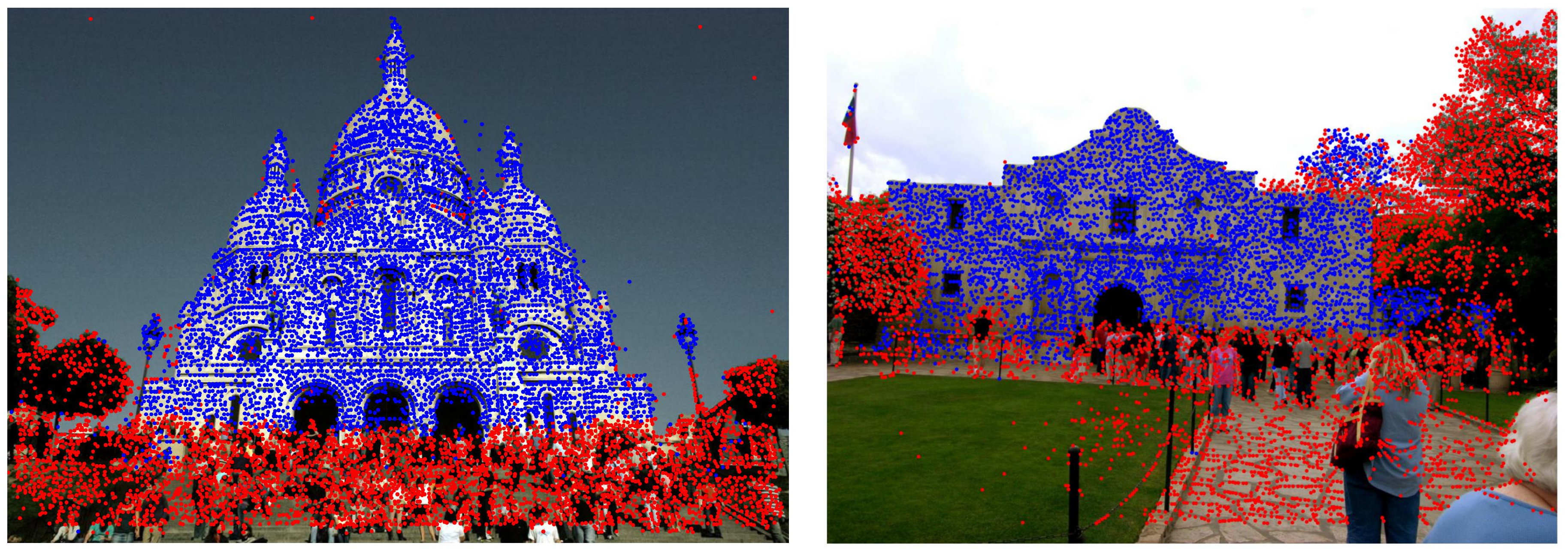}
	}
    \caption{An illustration of the influence of ambiguity for weakly supervised local feature methods. Keypoints that \textcolor{blue}{succeed} and \textcolor{red}{fail} to create landmarks are shown. (a) With \textit{\textbf{joint} training describe-then-detect} pipeline, DISK-W \cite{tyszkiewiczDISKLearningLocal2020a}  produces inaccurate keypoints that are out of the  objects. (b) With our \textit{\textbf{decoupled} training describe-then-detect} pipeline, PoSFeat can produce more reasonable keypoints. Best viewed in color.}
    \label{fig_ambiguity}
\end{figure}

Finding pixel correspondences is a fundamental problem in computer vision. {Sparse local feature \cite{lowe2004distinctive,bay2006surf,rublee2011orb,detone2018superpoint, wang2017persistence}, as one of the mainstream methods to find correspondences, has been widely applied in many areas,} such as simultaneous localization and mapping (SLAM) \cite{mur2017orb,zhao2019gslam}, structure from motion (SfM) \cite{schonberger2016structure,agarwal2011building}, and visual localization \cite{sattler2012improving, cai2019ground, zhang2021reference, guo2022soft}.

Traditional sparse local feature methods \cite{lowe2004distinctive, bay2006surf, rublee2011orb} follow a \textit{detect-then-describe}
pipeline. Specifically, keypoints are first detected and then patches centered at these keypoints are used to generate descriptors. 
{Early methods \cite{gunn1998edge, harris1988combined, lowe2004distinctive, leutenegger2011brisk} focus on the detection step and are proposed to distinguish distinctive areas to detect good keypoints. Later works pay more attention to the description step and make attempts to design powerful descriptors using advanced representations \cite{bay2006surf,calonder2011brief,rublee2011orb}.}

Motivated by the success of deep learning, many efforts \cite{yi2016lift,NIPS2017_831caa1b,ono2018lf,luo2019contextdesc,tian2019sosnet} have been made to replace the detection or description step in the \textit{detect-then-describe}
pipeline with CNNs. Recent works \cite{Dusmanu2019CVPR,revaud2019r2d2,luo2020aslfeat,tyszkiewiczDISKLearningLocal2020a} find that keypoints and descriptors are interdependent and propose a \textit{\textbf{joint} training describe-then-detect} pipeline. Specifically, the description network and detection network are combined into a single CNN and optimized jointly. The \textit{\textbf{joint} training describe-then-detect} pipeline achieves better performance than the \textit{detect-then-describe} pipeline, especially under challenging conditions \cite{toft2020long, jin2021image}. However, these methods are fully supervised and rely on dense ground-truth correspondence labels for training.

Because collecting a large dataset with  pixel-level ground-truth correspondences is expensive, self-supervised and weakly supervised learning are investigated for training. Specifically, DeTone \etal \cite{detone2018superpoint} used a single image and a virtual homography to generate image pairs to conduct self-supervised learning. However, homography transformation cannot cover complicated geometry transformations in real-world settings, resulting in limited performance. Noh \etal~\cite{noh2017large} used landmark labels to train the local feature network, which suffers extremely poor performance on viewpoint changes. Owing to the convenience of collecting camera poses, Wang \emph{et al.} \cite{wangLearningFeatureDescriptors2020} introduced camera poses as weak supervision for descriptor learning. Although weakly supervised learning achieves promising results within the \textit{detect-then-describe} pipeline, directly applying it to the \textit{\textbf{joint} training describe-then-detect} pipeline is hard to produce satisfying results \cite{tyszkiewiczDISKLearningLocal2020a}. 

\begin{figure}[t]
    \centering
    \includegraphics[width=0.45\textwidth]{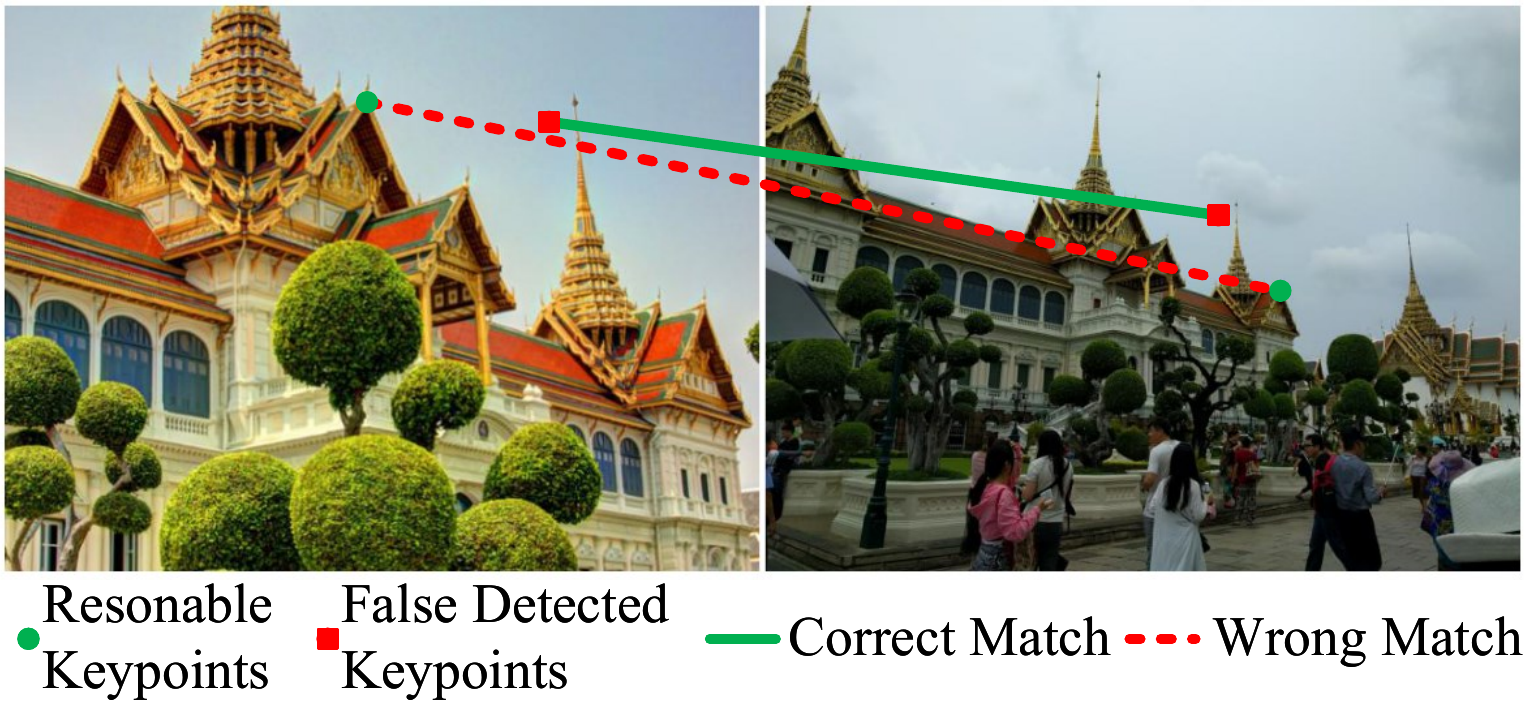}
    \caption{Motivation of decoupling. Two reasonable keypoints can be matched incorrectly due to the low discriminativeness of descriptors (e.g., caused by repetitive textures). Meanwhile, two false detected keypoints can also be matched with a high descriptor similarity. Best viewed in color.}
    \label{fig_intuition}
\end{figure}

When a detection network and a description network are jointly optimized within a \textit{\textbf{joint} training describe-then-detect} pipeline with only weak supervision (\emph{e.g.}, camera pose), the loss produced by these two components cannot be distinguished. Specifically, when only one component is failed (Fig. \ref{fig_intuition}), both the detection network and the description network cannot be correctly updated within a \textit{\textbf{joint} training describe-then-detect} pipeline. As a result, the description network is hard to produce highly discriminative descriptors, and the detection network may produce false detected keypoints that are out of object boundaries, as shown in Fig. \ref{fig_ambiguity}.

In this paper, we propose a \textit{\textbf{decoupled} training describe-then-detect} pipeline tailored for weakly supervised local feature learning. Our main insight is that, with only weak supervision, the detection network relies heavily on a good descriptor for accurate keypoint detection (Fig.~\ref{fig_ambiguity}). Consequently, we decouple the detection network from the description network to postpone it until a discriminative and robust descriptor is learned. Different from the \textit{detect-then-describe} pipeline that relies on low-level structures for early detection, our keypoints detection depends on the higher-level structures encoded in the descriptors. As a result, better robustness is achieved. In contrast to the \textit{\textbf{joint} training describe-then-detect} pipeline that simultaneously perform detection and description optimization, the two networks are trained separately and thus the loss function for these two components are decoupled to address the ambiguity. It is demonstrated that our \textit{\textbf{decoupled} training describe-then-detect} pipeline facilitates local feature methods to achieve much better performance with only weak supervision. Our contributions can be summarized as:

(1)~We introduce a \textit{\textbf{decoupled} training describe-then-detect} pipeline  for weakly supervised local feature learning. This simple yet efficient pipeline significantly improves the performance of weakly supervised local features. 

(2)~We propose a line-to-window search strategy to exploit the weak supervision of camera poses for descriptor learning. This strategy can make full use of the geometric information of camera poses to reduce the search space and learn highly discriminative descriptors.

{(3)~Our method achieves state-of-the-art performance on three datasets and largely closes the gap between fully and weakly supervised methods.}

\section{Related Works}

\begin{figure*}
    \centering
    \includegraphics[width=0.95\textwidth]{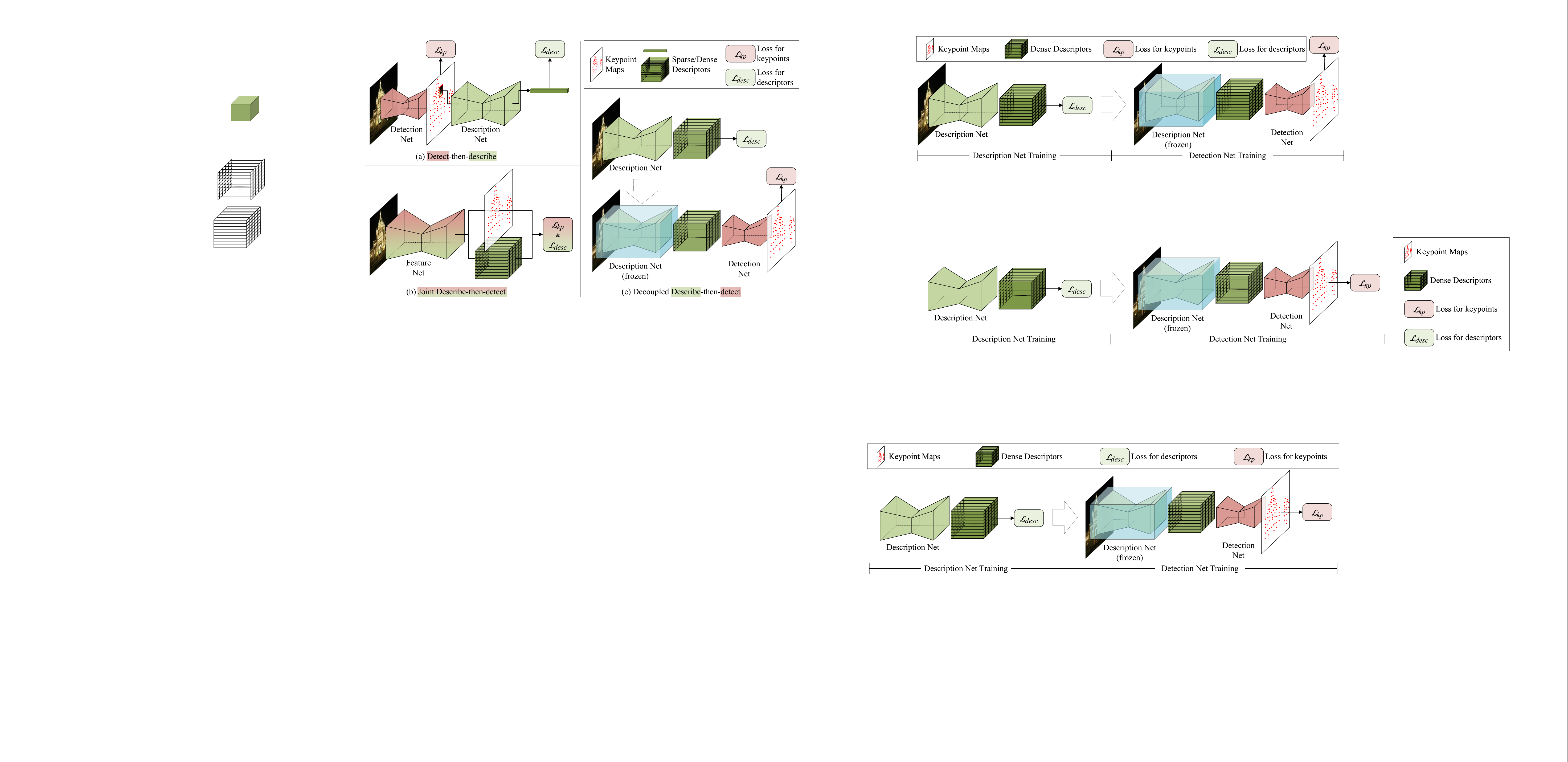}
    \caption{The proposed \textit{\textbf{decoupled} training describe-then-detect} pipeline. The detection network is decoupled from the description network and postponed until good descriptors are obtained.}
    \label{fig_pipeline}
\end{figure*}

\subsection{Fully Supervised Local Feature Methods}
Fully supervised methods conduct local feature learning using pixel-level ground-truth correspondences to provide supervision.
Following the \textit{detect-then-describe} pipeline, early learning-based methods \cite{savinov2017quad, barroso2019key,NIPS2017_831caa1b, tian2019sosnet,ebel2019beyond, luo2019contextdesc} use CNNs to perform the detection or description steps. Specifically, 
QuadNet \cite{savinov2017quad} and Key.Net \cite{barroso2019key} were proposed to use CNNs for keypoint detection. 
HardNet \cite{NIPS2017_831caa1b} and SOSNet \cite{tian2019sosnet} were developed to leverage CNNs to extract descriptors. 
Later, LIFT \cite{yi2016lift} and LFNet \cite{ono2018lf} were introduced to integrate both {detection} and {description} steps into an end-to-end architecture to achieve better performance. Note that, LIFT \cite{yi2016lift} also introduced a decoupled training to address unstable training issue with full supervision in a \textit{detect-then-describe} pipeline.

Recent works \cite{Dusmanu2019CVPR, revaud2019r2d2, luo2020aslfeat, tyszkiewiczDISKLearningLocal2020a} follow a \textit{\textbf{joint} training describe-then-detect} pipeline in which detection and description are combined into a single CNN and optimized jointly.
Specifically, Dusmanu \emph{et al.} \cite{Dusmanu2019CVPR} first used a CNN to extract dense features and then selected local maxima of the dense feature map as keypoints. Revaud \emph{et al.} \cite{revaud2019r2d2} further took  both the repeatablity and reliability of the descriptors into consideration for better keypoint detection. Tyszkiewicz \etal \cite{tyszkiewiczDISKLearningLocal2020a} used policy gradient to address the  discreteness during the selection of sparse keypoints (namely, DISK). Luo \emph{et al.} \cite{luo2020aslfeat} adopted deformable convolution to model the geometry information and detected keypoints at multiple scales.
By jointly optimizing the detection network and the description network, \textit{\textbf{joint} training describe-then-detect} pipeline achieves better performance than previous \textit{detect-then-describe} pipeline.

\subsection{Self-Supervised Local Feature Methods}
As a large dataset with densely labeled correspondences is difficult to collect, self-supervised learning has been studied for local feature learning. 
Specifically, DeTone \etal \cite{detone2018superpoint} used a virtual homography to generate an image pair from a single image to conduct self-supervised learning. 
This method uses a CNN pretrained on synthetic data as a teacher of the detection network. Differently, Christiansen \etal \cite{christiansen2019unsuperpoint}  proposed an end-to-end framework to train both the detection network and the description network using virtual homography in a self-supervised manner.
Later, Parihar \etal \cite{parihar2021rord}  leveraged the homography to enhance the robustness of descriptors to rotation.
Nevertheless, simple homography transformations used in these self-supervised methods may not hold in real cases. 

\subsection{Weakly Supervised Local Feature Methods}
Noh \etal introduced DELF \cite{noh2017large}, which is trained with an image retrieval task, to achieve local feature extraction. However, the keypoints detected by DELF are sensitive to vierwpoint changew and thus cannot be applied in real world settings. For camera poses are easy to collect, Wang \etal \cite{wangLearningFeatureDescriptors2020} used them as weak supervision and introduced an epipolar loss for descriptor learning. This method follows a \textit{detect-then-describe} pipeline and relies on an off-the-shelf detection method (\emph{e.g.}, SIFT) to detect keypoints.  
Recently, Tyszkiewicz \etal \cite{tyszkiewiczDISKLearningLocal2020a} developed DISK-W to integrate weakly supervised learning in a \textit{\textbf{joint} training describe-then-detect} pipeline by adopting policy gradient. 
Nevertheless, when DISK-W is directly trained with a weakly supervised loss (rather than a fully-supervised loss), it suffers a notable performance drop on pixel-wise metrics. As weakly supervised loss cannot distinguish between errors introduced by false keypoints and inaccurate descriptors, this ambiguity hinders the \textit{\textbf{joint} training describe-then-detect} pipeline to learn good local features. 

\subsection{Learning-based Matcher Methods}
\label{related_matcher}
Since a Brute Force Matcher (also named NN matcher) usually produces low quality raw matches, learning-based matchers are proposed to achieve better matching results. Sarlin \etal~\cite{sarlin2020superglue} proposed SuperGlue to achieve robust matching with a graph neural network (GNN) and an optimal transport algorithm. Chen \etal~\cite{chen2021learning} improved the architecture of GNN to increase the efficiency of descriptor enhancement. Zhou \etal~\cite{zhou2021patch2pix} proposed a weakly supervised network to refine raw matches using patch matches as prior. Sun \etal~\cite{sunLoFTRDetectorFreeLocal2021} introduced a detector-free matcher to achieve pixel correspondence in a coarse-to-fine manner. Note that, most matcher methods are not the direct competitors of local feature methods. Instead, they can be considered as a post processing step and combined with local features to achieve improved performance.

\begin{figure*}[t]
    \centering
    \includegraphics[width=0.85\textwidth]{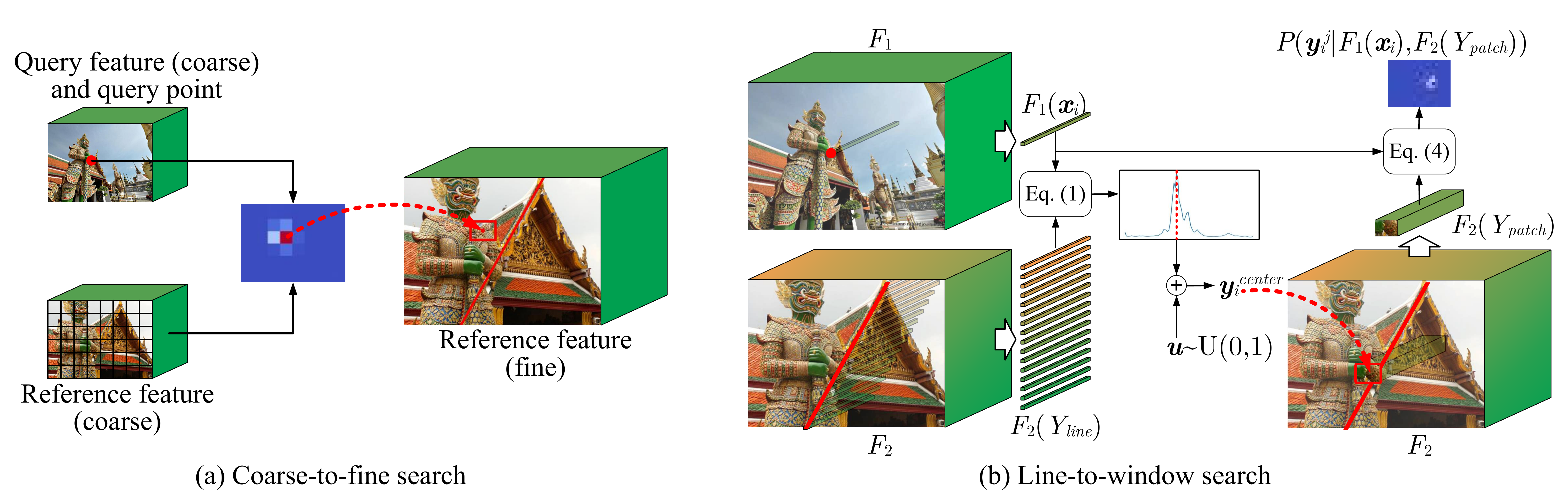}
    \caption{An illustration of the coarse-to-fine search strategy (a) and our line-to-window search strategy (b). The red line in $F_2$ denotes the epipolar line corresponding to the query point in $F_1$.}
    \label{fig_l2w}
\end{figure*}

\section{Decoupled Training Describe-then-Detect Pipeline}

\subsection{Overview}
\label{text_overview}
The \textit{\textbf{decoupled} training describe-then-detect} pipeline is shown in Fig.~\ref{fig_pipeline}. We train the description net and detection net individually to suppress the loss ambiguity caused by weak supervision.  During training, we first leave out the detection network and optimize the description network to learn good descriptors with a line-to-window strategy. The description network is then frozen to train a detection network for keypoint detection. We follow CAPS~\cite{wangLearningFeatureDescriptors2020} to use ResUNet as the description net, which produces a feature map with 1/4 resolution and 128 dimensions as dense descriptors. Additionally, we design a shallow detection net to detect keypoints at the original resolution. For more details about the network architecture, please refer to the supplementary material.

\subsection{Feature Description}
\label{text_method_desc}
%We adopt an encoder-decoder network as the descriptor to produce dense feature description. 
Following the widely used paradigm \cite{wangLearningFeatureDescriptors2020}, we impose supervision only on sparse query points sampled from paired images to conduct training of the description network. We first split an image into small grids of size $g_d\times g_d$, and randomly sample one point per grid as a query point. Then, we translate
relative camera pose into an epipolar constraint and  introduce a line-to-window search strategy to reduce search space  (Sec.~\ref{text_method_desc_l2w}). 
Moreover, we formulate a loss function by encouraging the predicted matches to obey the epipolar constraint  (Sec.~\ref{sec_loss}).

\subsubsection{Line-to-Window Search}
\label{text_method_desc_l2w}
Given a query point $\boldsymbol {x}$ in the query image $I_1$, our goal is to find its correspondence in the reference image $I_2$. 
Since repetitive structures widely exist in a natural image, the commonly used coarse-to-fine strategy \cite{wangLearningFeatureDescriptors2020, sunLoFTRDetectorFreeLocal2021} usually selects a mismatched patch such that inferior performance is produced (Fig.~\ref{fig_l2w}(a)). 
Intuitively, the correspondence of the query point $\boldsymbol{x}$ is constrained in an epipolar line in the reference image. Therefore, we introduce a line-to-window search strategy to reduce search space for better performance. Our line-to-window search strategy consists of two search steps, as illustrated in Fig.~\ref{fig_l2w}(b).

\noindent \textbf{Search along An Epipolar Line.}

For a query point $ \boldsymbol{x}_i \in I_1$, we first calculate its corresponding epipolar line $L_{\boldsymbol{x}_i}$ in the reference image $I_2$ based on the relative camera pose. Then, we uniformly sample $N_{line}$ points along this epipolar line to formulate the search space $Y_{line}=\{\boldsymbol{y}_i^j\}(j=1,...,N_{line})$.
Next, we calculate the matching probability of $\boldsymbol{x}_i$ over $Y_{line}$:
\begin{equation}
\label{eq_probfunc}
    P({\boldsymbol{y}_i^j}|F_1({\boldsymbol{x}_i}),F_2(Y_{line}))=\frac{\exp (F_1({\boldsymbol{x}_i})^{\rm T}F_2({\boldsymbol{y}_i^j}))}{\sum_{ Y_{line}}\exp (F_1({\boldsymbol{x}_i})^{\rm T}F_2({\boldsymbol{y}_i^k}))},
\end{equation} 
where $F_1$ and $F_2$ are the feature maps for $I_1$ and $I_2$, respectively. Afterwards, we select $\overline{\boldsymbol{y}}_i$ with the maximum probability from $Y_{line}$ to determine the coarse location of the correspondence of $\boldsymbol{x}_i$: 
\begin{equation}
\label{eq_dmfunc}
    \overline{\boldsymbol{y}}_i=\mathop{\arg\max}\limits_{ {\boldsymbol{y}_i^j}}P({\boldsymbol{y}_i^j}|F_1( {\boldsymbol{x}_i}),F_2(Y_{line})).
\end{equation}

\noindent \textbf{Search in A Local Window.}

Due to the discreteness of the candidates in $Y_{line}$, the resultant corresponding point $\overline{\boldsymbol{y}}_i$ can be far from the groundtruth. To remedy this, a subsequent search is conducted in a local window. First, we calculate the center of the local window:
\begin{equation}
\label{eq_l2w}
     \boldsymbol{y}_i^{center} = \overline{\boldsymbol{y}}_i+0.5\cdot w_{patch}\cdot  \boldsymbol{u},
\end{equation}
where $w_{patch}$ is the window size of a local patch, $ \boldsymbol{u}\in\mathbb{R}^2$ is a noise vector drawn from a uniform distribution ${\rm U}(0,1)$ to avoid the convergence to trivial solution $F( \boldsymbol{x})\equiv{0}$.
Then, a local patch $Y_{patch}\subset I_2$ centered at $\boldsymbol{y}_i^{center}$ is cropped from $F_2$ as the search space. Next, we calculate the matching probability of $\boldsymbol{x}_i$ over $Y_{patch}$:
\begin{equation}
\label{eq_probfunc_patch}
    P({\boldsymbol{y}_i^j}|F_1({\boldsymbol{x}_i}),F_2(Y_{patch}))=\frac{\exp (F_1({\boldsymbol{x}_i})^{\rm T}F_2({\boldsymbol{y}_i^j}))}{\sum_{Y_{patch}}\exp (F_1({\boldsymbol{x}_i})^{\rm T}F_2({\boldsymbol{y}_i^k}))}.
\end{equation} 
Because directly selecting the point with the maximum probability in the local patch is non-differentiable, we calculate the correspondence $\hat{\boldsymbol{y}}_i$ in a differentiable manner:
\begin{equation}
\label{eq_dmfunc_patch}
    \hat{\boldsymbol{y}}_i=E({\boldsymbol{y}_i^j})=\sum_{{\boldsymbol{y}_i^j}\in Y_{patch}}{\boldsymbol{y}_i^j}\cdot P({\boldsymbol{y}_i^j}|F_1({\boldsymbol{x}_i}),F_2(Y_{patch})).
\end{equation}

Compared to the previous coarse-to-fine search strategy \cite{wangLearningFeatureDescriptors2020}, our line-to-window search strategy can make better use of the camera pose information to reduce search space and further improve the discriminativeness of descriptors (as demonstrated in Sec.~\ref{ablation_2stage}).

\subsubsection{Loss Function}
\label{sec_loss}
With only weak supervision of camera pose, we calculate the distance of the correspondence $\hat{\boldsymbol{y}}_i$ to the epipolar line $L_{\boldsymbol{x}_i}$ as the loss of query point $\boldsymbol{x}_i$   \cite{wangLearningFeatureDescriptors2020}:
\begin{equation}
    \label{eq_epiloss}
    \mathcal{L}_{epi}( {\hat{\boldsymbol{y}}_i},  \boldsymbol{x}_i)={\rm distance}({\hat{\boldsymbol{y}}_i},~{L_{\boldsymbol{x}_i}}).
\end{equation}
Then, we use the weighted sum of the losses over all query points as the final loss:
\begin{equation}
    \mathcal{L}_{desc}= \frac{\sum_{i} \frac{M_i}{ {\sigma}( \boldsymbol{x}_i)} \cdot\mathcal{L}_{epi}( \hat{ \boldsymbol{y}}_{i},  {\boldsymbol{x}_i})}{\sum_i \frac{M_i}{ {\sigma}( \boldsymbol{x}_i)}}.
\end{equation}
Here, $M_i$ is a binary mask (which is used to exclude query points whose epipolar lines are not in the reference image) and  $ {\sigma}( \boldsymbol{x}_i)$ is the variance of the probability distribution over $Y_{patch}$,
\begin{equation}
    {\sigma}( \boldsymbol{x}_i)=\lVert \hat{\boldsymbol{y}}_i^2 - E({\boldsymbol{y}_i^{j 2}}) \rVert
\end{equation}

\subsection{Feature Detection}
\label{text_method_dete}
After feature description learning, the description network is frozen to produce dense descriptors for keypoint detection, as shown in Fig.~\ref{fig_pipeline}. 
Since selecting discrete sparse keypoints is non-differentiable, we adopt the strategy introduced in DISK \cite{tyszkiewiczDISKLearningLocal2020a}, which is  based on policy gradient, to achieve network training.

First, dense descriptors $F_1$ and $F_2$ are respectively extracted from $I_1$ and $I_2$, and fed to a detection network to produce keypoint heatmaps. Then, we divide these heatmaps into grids of size $g_k\times g_k$ and select at most one keypoint from each grid cell. Specifically, we establish a probability distribution $P_{kp}$ over each grid cell based on the heatmap scores in this cell. Afterwards, $P_{kp}$ is used to probabilistically select candidate keypoints $Q_1\!=\!\{ {\boldsymbol{x}_1}, {\boldsymbol{x}_2},\cdots\}$ and $Q_2\!=\!\{ {\boldsymbol{y}_1}, {\boldsymbol{y}_2},\cdots\}$ from $I_1$ and $I_2$, respectively. Next, a matching probability $P_m$ is calculated based on the feature similarity $S_{i,j}$ between each pair of candidate keypoints $(\boldsymbol{x}_i,\boldsymbol{y}_j)$. 
With only camera pose supervision, we adopt an epipolar reward similar to Eq.~\ref{eq_epiloss} to encourage $\boldsymbol{y}_j$ to be close to the epipolar line of $\boldsymbol{x}_i$ (\emph{i.e.}, $L_{\boldsymbol{x}_i}$):
\begin{equation}
    R( \boldsymbol{x}_i, \boldsymbol{y}_j) = 
    \begin{cases} 
    \lambda_p,  & \mbox{if }{\rm distance}( \boldsymbol{y}_j, L_{\boldsymbol{x}_i}) \leq\epsilon \\
    \lambda_n, & \mbox{if }{\rm distance}( \boldsymbol{y}_j,  L_{\boldsymbol{x}_i}) >\epsilon
    \end{cases},
\end{equation}
where the reward threshold $\epsilon$ is empirically set to 2. The overall loss function is defined as: 
\begin{equation}
\begin{split}
    \mathcal{L}_{kp}= &-\frac{1}{|Q_1|+|Q_2|} \Big (\sum_{ \boldsymbol{x}_i, \boldsymbol{y}_j}\mathcal{L}_{rew}( \boldsymbol{x}_i, \boldsymbol{y}_j)\\
    &+ \lambda_{reg}\big(\sum_{\boldsymbol{x}_i} \log P_{kp}( \boldsymbol{x}_i) + \sum_{\boldsymbol{y}_j} \log P_{kp}( \boldsymbol{y}_j)\big) \Big ),
\end{split}
\end{equation}
where $\lambda_{reg}$ is a regularization penalty and the reward loss $\mathcal{L}_{rew}( \boldsymbol{x}_i, \boldsymbol{y}_j)$ is defined as:
\begin{equation}
    \mathcal{L}_{rew}( \boldsymbol{x}_i, \boldsymbol{y}_j) \!=\! P_{m}( \boldsymbol{x}_i, \boldsymbol{y}_j)\cdot R( \boldsymbol{x}_i, \boldsymbol{y}_j) \cdot \log (P_{kp}( \boldsymbol{x}_i)P_{kp}( \boldsymbol{y}_i)).
%         &\cdot (\log P( {x}_i) + \log P( {y}_j))
\end{equation}
Since our descriptors are well optimized, $P_{m}$ can suppress spurious points with low scores. In contrast, in a joint pipeline,  descriptors are under-optimized such that spurious points cannot be well distinguished. Please refer to the supplementary material for more details.

\section{Experiments}
\subsection{Experimental Settings}
\paragraph{Datasets} The MegaDepth dataset \cite{li2018megadepth} was used for training. We used a subset of the training split of CAPS \cite{wangLearningFeatureDescriptors2020}. Totally, 127 out of 196 scenes were used as the training set. 

\noindent\textbf{Implementation Details} 
During the training phase, images were resized to $640\times480$ with breaking the aspect ratio. All networks were trained using a SGD optimizer with nesterov momentum \cite{sutskever2013importance}. The learning rate is set to $1\times 10^{-3}$ and the batch size was set to 6. The description network was trained for 100,000 iterations, and the detection network was trained for 5,000 iterations. All experiments were conducted using Pytorch on a single NVIDIA RTX3090 GPU.
In our experiments, the number of sampled points $N_{line}$ was set to 100, the window size $w_{patch}$ was set to 0.1 (normalized height and width), and the grid size $g_d$ and $g_k$ were set to 16 and 8, respectively. Following \cite{tyszkiewiczDISKLearningLocal2020a}, $\lambda_{p}$, $\lambda_{n}$, and $\lambda_{reg}$ were set to 1, -0.25, and -0.001,  respectively. For more details, please refer to the supplementary material.

\subsection{Comparison with Previous Methods}
\begin{figure*}[!t]
    \centering
    \includegraphics[width=0.95\textwidth]{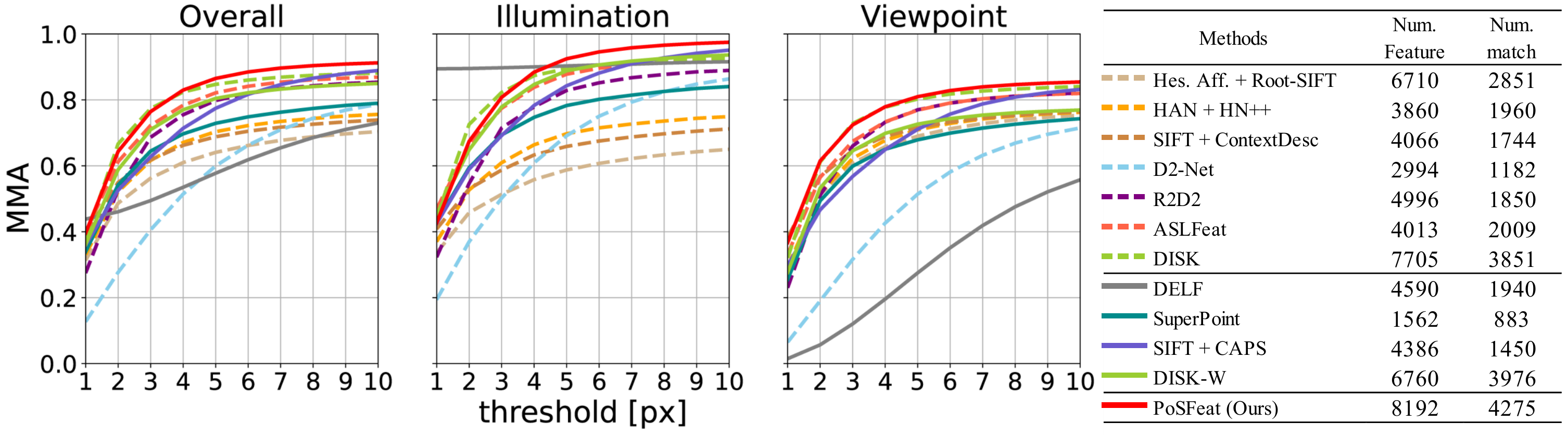}
    \caption{Results achieved on the HPatches dataset \cite{hpatches_2017_cvpr}. Mean match accuracy (MMA) achieved at different thresholds are illustrated. Learning based methods with weak supervision are shown in solid lines while other methods are shown in dashed lines. The numbers of keypoints and matches for each method are also reported.}
    \label{fig-reshpatches}
\end{figure*}

\begin{figure*}[!t]
    \centering
    \includegraphics[width=0.9\textwidth]{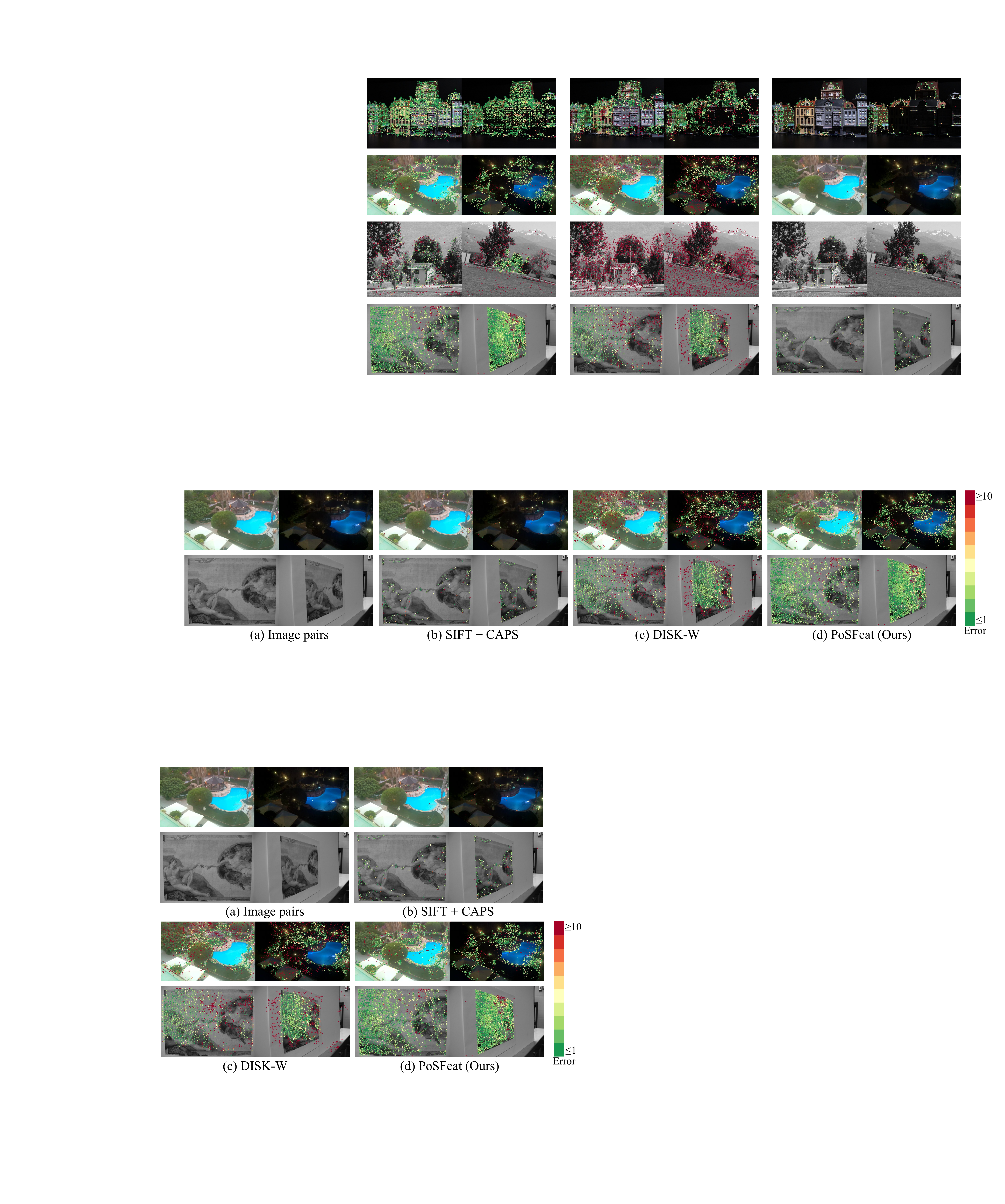}
    \caption{Visualization results achieved on HPatches. For simplicity,  only successfully matched keypoints are shown and colored according to their match errors. The colorbar is shown on the right. Best viewed in color.}
    \label{fig_vishpatches}
\end{figure*}

\subsubsection{Feature Matching}
\textbf{Settings}. We first evaluate our method on the widely used HPatches dataset \cite{hpatches_2017_cvpr}.Following D2-Net \cite{Dusmanu2019CVPR}, 8 high-resolution scenes are removed and the remaining 52 scenes with illumination changes and 56 scenes with viewpoint changes are included for evaluation. Mean matching accuracy (MMA) \cite{Dusmanu2019CVPR} with thresholds ranging from 1 to 10 is used for evaluation. 
We also use a weighted sum of MMA at different thresholds for overall evaluation:

\begin{equation}
\label{eq_mmascore}
    {\rm MMAscore} =  \frac{\sum_{{\rm thr}\in [1,10]}(2-0.1\cdot {\rm thr})\cdot {\rm MMA@thr}}{\sum_{{\rm thr}\in [1,10]}(2-0.1\cdot {\rm thr})}.
\end{equation}

{Three families of methods are included for comparison:}
\begin{itemize}
    \item \textbf{{Patch-based methods:}} Hessian-Affine keypoints \cite{mikolajczyk2004scale} with Root-SIFT \cite{arandjelovic2012three} (Hes. Aff. + Root-SIFT), affine region detector HesAffNet \cite{mishkin2018repeatability} with HardNet++ \cite{NIPS2017_831caa1b} (HAN + HN++), and SIFT \cite{lowe2004distinctive} with ContextDesc \cite{luo2019contextdesc} (SIFT + ContextDesc).

    \item { \textbf{Fully supervised dense feature methods:}} D2-Net \cite{Dusmanu2019CVPR}, R2D2 \cite{revaud2019r2d2}, ASLFeat \cite{luo2020aslfeat}, and DISK \cite{tyszkiewiczDISKLearningLocal2020a}. 

    \item { \textbf{Weakly supervised dense feature methods:}} DELF \cite{noh2017large}, SuperPoint \cite{detone2018superpoint}, DISK-W \cite{tyszkiewiczDISKLearningLocal2020a}, 
    and SIFT with CAPS \cite{wangLearningFeatureDescriptors2020} (SIFT + CAPS).
\end{itemize}

\begin{table}[t]
    \centering
    \small
    \resizebox{0.48\textwidth}{!}{
    \begin{tabular}{l|c c c}
        \hline
         Methods & \makecell[c]{\footnotesize{MMAscore}\\\footnotesize{Overall}} & \makecell[c]{\footnotesize{MMAscore}\\ \footnotesize{Illumination}} & \makecell[c]{\footnotesize{MMAscore}\\ \footnotesize{Viewpoint}}  \\
         \hline
         \footnotesize{Hes. Aff. + Root-SIFT}~\cite{arandjelovic2012three} & 0.584 & 0.544 & 0.624\\
         \footnotesize{HAN ~\cite{mishkin2018repeatability} + HN++~\cite{NIPS2017_831caa1b}} & 0.633 & 0.634 & 0.633\\
         SIFT~\cite{lowe2004distinctive} + ContextDesc~\cite{luo2019contextdesc} & 0.636 & 0.613 & 0.657\\
         D2Net~\cite{Dusmanu2019CVPR} & 0.519 & 0.605 & 0.440\\
         R2D2~\cite{revaud2019r2d2} & 0.695 & 0.727 & 0.665\\
         ASLFeat~\cite{luo2020aslfeat} & 0.739 & 0.795 & 0.687\\
         DISK~\cite{tyszkiewiczDISKLearningLocal2020a} & \underline{0.763} & 0.813 & \underline{0.716}\\
         \hline 
         DELF~\cite{noh2017large} & 0.571  & \textbf{0.903} & 0.262\\
         SuperPoint~\cite{detone2018superpoint} & 0.658 & 0.715 & 0.606\\
         SIFT~\cite{lowe2004distinctive} + CAPS~\cite{wangLearningFeatureDescriptors2020} & 0.699 & 0.764 & 0.639\\
         DISK-W~\cite{tyszkiewiczDISKLearningLocal2020a} & 0.719 & 0.803 & 0.649\\
         \hline 
         PoSFeat (Ours) & \textbf{0.775} & \underline{0.826} & \textbf{0.728}\\
         \hline
    \end{tabular}
    }
    \caption{MMAscore results achieved by different methods on the HPatches dataset \cite{hpatches_2017_cvpr}. The MMAscores are calculated from Fig.~\ref{fig-reshpatches}.}
    \label{tab_reshpatches}
\end{table}

\noindent\textbf{Results}. 
As shown in Fig.~\ref{fig-reshpatches} and Table~\ref{tab_reshpatches}, the proposed PoSFeat outperforms all previous works, with the highest MMAscore being achieved. Compared to existing weakly supervised methods, our method produces significant performance improvements. Specifically, our method outperforms DISK-W by notable margins under both illumination (0.826 vs. 0.803) and viewpoint (0.728 vs. 0.649) changes, and therefore achieves higher overall MMAscore (0.775 vs. 0.719). We also visualize the matching results in Fig. \ref{fig_vishpatches}. It can be seen that our PoSFeat produces more reasonable keypoints and less wrong matches. Compared to fully supervised methods, our method still performs favorably with higher MMA scores. This clearly demonstrates the superiority of our method. Note that, because DELF detects keypoints in a low resolution feature map with a fixed grid, it produces the best results under illumination change. However, our method significantly surpasses DELF under viewpoint change (0.728 vs. 0.262) and achieves much better overall performance (0.775 vs. 0.571).

\subsubsection{Visual Localization}
\textbf{Settings}. We then evaluate our method on the visual localization task with the Aachen Day-Night dataset \cite{zhang2021reference}.We adopt the official visual localization pipeline\footnote{https://github.com/tsattler/visuallocalizationbenchmark/tree/master/ local\_feature\_evaluation} used in the local feature challenge of workshop on long-term visual localization under changing conditions. This challenge only evaluates the pose of night-time query images. Accuracy with different thresholds are used as metrics, including (0.5m, 2$^\circ$), (1m, 5$^\circ$), and (5m, 10$^\circ$).

We compare our method with two families of methods:

\begin{itemize}
    \item \textbf{Local feature methods:} D2-Net \cite{Dusmanu2019CVPR}, SuperPoint \cite{detone2018superpoint}, R2D2 \cite{revaud2019r2d2}, ASLFeat \cite{luo2020aslfeat}, ISRF \cite{melekhov2020image}, and LISRD \cite{pautrat2020online}. 
    \vspace{-0.25cm}
    
    \item \textbf{Matcher methods:}  DualRC-Net \cite{li2020dual}, SuperGlue \cite{sarlin2020superglue} + SuperPoint, SparseNCNet~\cite{rocco2020efficient}, LoFTR \cite{sunLoFTRDetectorFreeLocal2021}, Patch2Pix~\cite{zhou2021patch2pix}, and SGMNet \cite{chen2021learning} + SuperPoint. As mentioned in Sec~\ref{related_matcher}, matchers are the cooperators instead of the direct competitors of local features. Therefore, we group them separately.
\end{itemize}

\begin{table}[t]
    \centering
    \small
    \resizebox{0.48\textwidth}{!}{
        \begin{tabular}{l|c c c |c c c}
        \hline
            \multirow{2}{*}{Method} & \multicolumn{3}{c|}{ Aachen Day-Night v1}&\multicolumn{3}{c}{Aachen Day-Night v1.1} \\
            \cline{2-7}
             ~ & (0.5m,2$^\circ$) & (1m,5$^\circ$) & (5m, 10$^\circ$) & (0.5m,2$^\circ$) & (1m,5$^\circ$) & (5m, 10$^\circ$)\\
        \hline 
         SP \cite{detone2018superpoint} & 74.5 & 78.6 & 89.8 & - & - & - \\
         D2-Net \cite{Dusmanu2019CVPR} & 74.5 & 86.7 & \textbf{100} & - & - & -\\
         R2D2 \cite{revaud2019r2d2} & 76.5 & \textbf{90.8} & \textbf{100} & {71.2} & 86.9 & 97.9\\
         ASLFeat \cite{luo2020aslfeat} & \textbf{81.6} & 87.8 & \textbf{100} & - & - & -\\
         ISRF \cite{melekhov2020image} & - & - & - & 69.1 & \textbf{87.4} & \textbf{98.4} \\
         LISRD \cite{pautrat2020online}& - & - & - & \underline{73.3} & 86.9 & 97.9 \\
         PoSFeat (Ours) & \textbf{81.6} & \textbf{90.8} & \textbf{100} & \textbf{73.8} & \textbf{87.4} & \textbf{98.4}\\
        \hline 
        \hline
         DualRC-Net \cite{li2020dual} & - & - & - & 71.2 & 86.9 & 97.9\\
         SP+SuperGlue \cite{sarlin2020superglue} & 79.6 & {90.8} & {100} & 73.3 & 88.0 & 98.4\\
         Sparse-NCNet \cite{rocco2020efficient} & 76.5 & 84.7 & 98.0  & - & - & -\\
         LoFTR \cite{sunLoFTRDetectorFreeLocal2021} & - & - & - & 72.8 & {88.5} & {99.0}\\
         Patch2Pix \cite{zhou2021patch2pix} & 79.6 & 87.8 & 100 & - & - & -\\
         SP+SGMNet \cite{chen2021learning} & 77.6 & 88.8 & 99.0 & 72.3 & 85.3 & 97.9 \\
        \hline 
        \end{tabular}
    }
    \caption{Results achieved by different methods on the Aachen Day-Night dataset \cite{zhang2021reference}. `LISRD' represents LISRD with SuperPoint keypoints and AdaLAM \cite{cavalli2020handcrafted}. Two categories of methods are presented, including feature methods (top) and matchers (bottom).}
    \label{tab_vis_loc}
\end{table}

\noindent\textbf{Results}. As shown in Table \ref{tab_vis_loc}, our PoSFeat achieves the state-of-the-art performance among the feature methods. Specifically, on Aachen Day-Night v1, our method achieves the best accuracy in terms of all metrics. Note that, although ASLFeat is a fully supervised method, our PoSFeat still outperforms it on (1m, 5$^\circ$). On Aachen Day-Night v1.1, our method also produces the best performance in all metrics. Note that, although R2D2 \cite{revaud2019r2d2}, ISRF \cite{melekhov2020image}, and LISRD \cite{pautrat2020online} are fully-supervised and trained on the Aachen Day-Night dataset, our PoSFeat still achieves better results. We additionally include  matcher methods for further comparison. Although these methods take pairs of images as inputs, our PoSFeat achieves comparable or even better performance.

\begin{table}[t]
    \centering
    \footnotesize 
    \resizebox{0.45\textwidth}{!}{
    \begin{tabular}{c|l|c c c c}
    \hline
    Subset & Method & \makecell[c]{\# Imgs} & \makecell[c]{\# Pts} & \makecell[c]{Track \\Length} & \makecell[c]{Reproj. \\Err. (px)} \\
    \hline
         \multirow{5}*{\makecell*[l]{South \\Building \\(128 imgs)}} & Root-SIFT \cite{arandjelovic2012three, lowe2004distinctive}  & 128 & 108k & 6.32 & \textbf{0.55}\\
         ~ & SuperPoint \cite{detone2018superpoint}  & 128 & \textbf{160k} & 7.83 & 0.92 \\
         ~ & RFP \cite{bhowmik2020reinforced}  & 128 & 102k & 7.86 & 0.88 \\
         ~ & DISK \cite{tyszkiewiczDISKLearningLocal2020a}  & 128 & 115k & \textbf{9.91} & 0.59 \\
         ~ & DISK-W \cite{tyszkiewiczDISKLearningLocal2020a} & 128 & \underline{154k} & \underline{9.63} & 0.63 \\
         ~ & PoSFeat (Ours) & 128 & {148k} & {9.47} & \underline{0.58} \\
         \hline 
         \multirow{7}*{\makecell*[l]{Madrid \\Metropolis \\(1344 imgs)}} & Root-SIFT \cite{arandjelovic2012three, lowe2004distinctive}  & 500 & 116k & 6.32 & \textbf{0.60}\\
         ~ & SuperPoint \cite{detone2018superpoint} & 438 & 29k & \underline{9.03} & 1.02 \\
         ~ & D2-Net \cite{Dusmanu2019CVPR} & 501 & 84k & 6.33 & 1.28 \\
         ~ & ASLFeat \cite{luo2020aslfeat} & 613 & 96k & 8.76 & 0.90 \\
         ~ & CAPS \cite{wangLearningFeatureDescriptors2020} & \textbf{851} & \underline{242k} & 6.16 & 1.03 \\
         ~ & CoAM \cite{wiles2021coam} & \underline{702} & \textbf{256k} & 6.09 & 1.30 \\
         ~ & PoSFeat (Ours) & 419 & 72k & \textbf{9.18} & \underline{0.86} \\  
         \hline 
         \multirow{7}*{\makecell*[l]{Gendar-\\menmarkt \\(1463 imgs)}} & Root-SIFT \cite{arandjelovic2012three, lowe2004distinctive}   & 1035 & 339k & 5.52 & \textbf{0.70}\\
         ~ & SuperPoint \cite{detone2018superpoint}  & 967 & 93k & 7.22 & 1.03 \\
         ~ & D2-Net \cite{Dusmanu2019CVPR}  & 1053 & 250k & 5.08 & 1.19 \\
         ~ & ASLFeat \cite{luo2020aslfeat}  & 1040 & 221k & \textbf{8.72} & 1.00 \\
         ~ & CAPS \cite{wangLearningFeatureDescriptors2020}  & \textbf{1179} & \textbf{627k} & 5.31 & 1.00\\
         ~ & CoAM \cite{wiles2021coam} & \underline{1072} & \underline{570k} & 6.60 & 1.34\\
         ~ & PoSFeat (Ours) & 956 & 240k & \underline{8.40} & \underline{0.92} \\ 
         \hline 
         \multirow{7}*{\makecell*[l]{Tower of \\London \\(1576 imgs)}} & Root-SIFT \cite{arandjelovic2012three, lowe2004distinctive}  & 806 & 239k & 7.76 & \textbf{0.61}\\
         ~ & SuperPoint \cite{detone2018superpoint}& 681 & 52k & 8.67 & 0.96 \\
         ~ & D2-Net \cite{Dusmanu2019CVPR} & 785 & 180k & 5.32 & 1.24 \\
         ~ & ASLFeat \cite{luo2020aslfeat} & \underline{821} & 222k & \textbf{12.52} & 0.92 \\
         ~ & CAPS \cite{wangLearningFeatureDescriptors2020} & \textbf{1104} & \textbf{452k} & 5.81 & 0.98\\
         ~ & CoAM \cite{wiles2021coam} & 804 & 239k & 5.82 & 1.32 \\
         ~ & PoSFeat (Ours) & 778 & \underline{262k} & \underline{11.64} & \underline{0.90} \\
        \hline
    \end{tabular}
    }
    \caption{Results achieved by different methods on the ETH local feature benchmark.}
    \label{tab_3d_recon}
\end{table}

\subsubsection{3D Reconstruction}
\textbf{Settings}. We finally evaluate our method on the 3D reconstruction task. We conduct experiments on the ETH local feature benchmark \cite{schonberger2017comparative}. Four metrics are used for evaluation, including the number of registered images (\# Imgs), the number of sparse points (\# Pts), track length, and the mean reprojection error (Reproj. Err.).

Four families of methods were included for comparison:
\begin{itemize}
    \item \textbf{ Patch-based method:} Root-SIFT \cite{arandjelovic2012three, lowe2004distinctive}.
    \vspace{-0.25cm}
    
    \item\textbf{ Fully supervised dense feature methods:}  Reinforced Feature Points \cite{bhowmik2020reinforced} (RFP), DISK \cite{tyszkiewiczDISKLearningLocal2020a}, DISK-W \cite{tyszkiewiczDISKLearningLocal2020a}, D2-Net \cite{Dusmanu2019CVPR}, and ASLFeat \cite{luo2020aslfeat}. \vspace{-0.25cm}

    \item\textbf{ Weakly supervised dense feature methods:}  SuperPoint \cite{detone2018superpoint} and CAPS \cite{wangLearningFeatureDescriptors2020}. 
    \vspace{-0.25cm}

    \item\textbf{ Fully supervised matcher method:} CoAM \cite{wiles2021coam}.
    
\end{itemize}
 
\begin{figure}[t]
    \centering
    \includegraphics[width=0.5\textwidth]{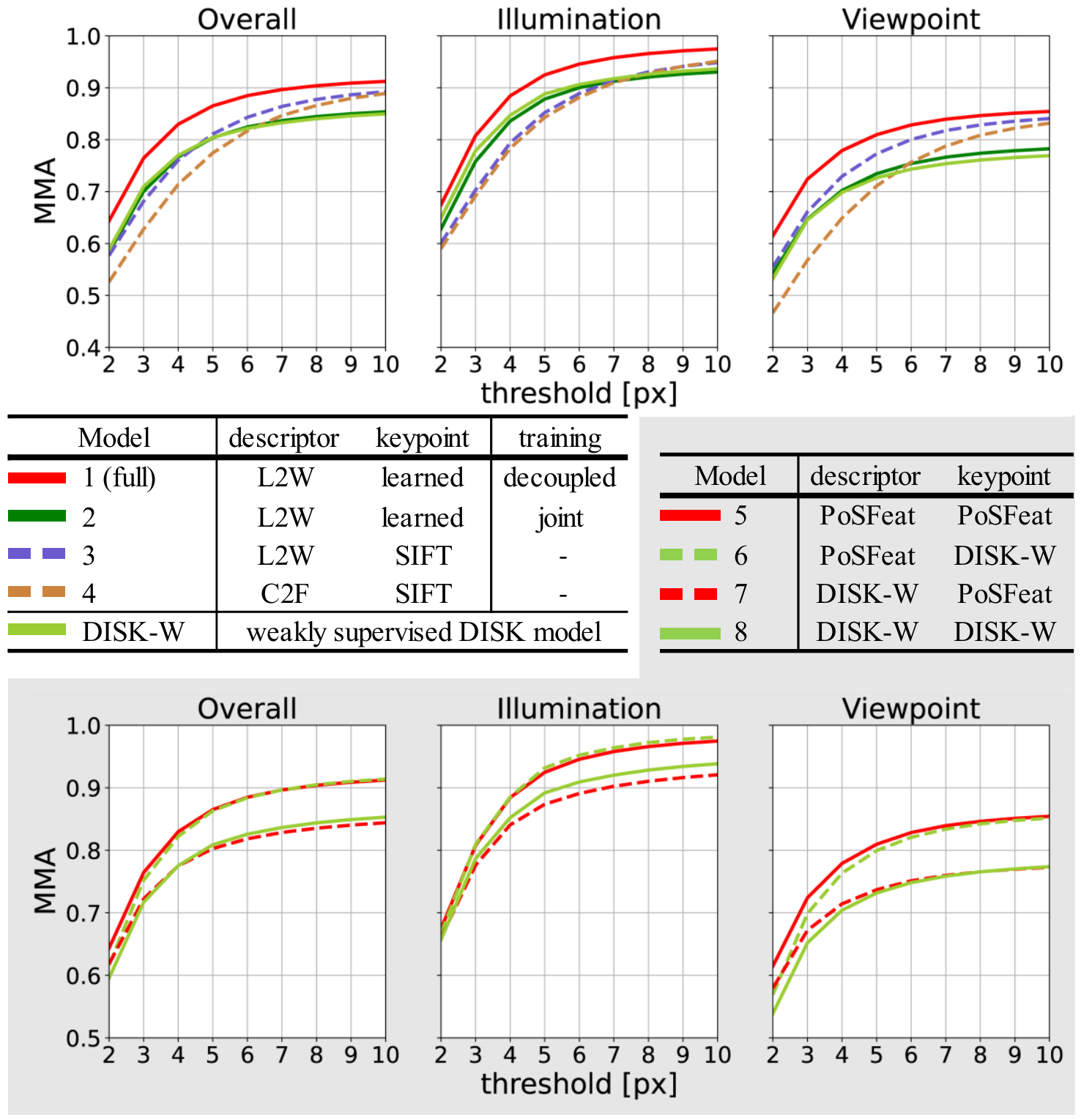}
    \caption{Ablation results on HPatches. ``L2W'' denotes our line-to-window search strategy (illustrated in Fig~\ref{fig_l2w}(b)) and ``C2F'' denotes the coarse-to-fine search strategy  \cite{wangLearningFeatureDescriptors2020} (illustrated in Fig~\ref{fig_l2w}(a)). ``learned'' means that the keypoints are generated by a detection network and ``SIFT'' mean that SIFT keypoints (OpenCV default settings) are used. ``decoupled'' means the proposed decoupled training pipeline is adopted and `joint' means the description network and the detection network are jointly optimized.}
    \label{fig_ablation_hpatch}
\end{figure}

\noindent\textbf{Results.} As shown in Table~\ref{tab_3d_recon}, our method performs favorably against previous methods on the 3D reconstruction task. Specifically, our method produces the lowest reprojection error among all learning-based methods.
Moreover, our method achieves the best or second best performance in terms of track length, which demonstrates that our keypoints are robust and thus can be tracked across a large amount of images.

\subsection{Ablation Study}
In this section, we first conduct ablation experiments on the HPatches dataset  \cite{hpatches_2017_cvpr}  to demonstrate the effectiveness of our \textit{\textbf{decoupled} training describe-then-detect} pipeline and line-to-window search strategy. Then, we conduct experiments to study the effectiveness of hyper-parameters in our method, \ie, the number of points sampled from the epipolar line $N_{line}$ and the window size $w_{patch}$. Results and model settings are shown in Fig. \ref{fig_ablation_hpatch} and Table \ref{tab_ablation_hyper}. 

\noindent\textbf{\textbf{Decoupled} Training Describe-then-Detect Pipeline.}
\label{ablation_2stage}
We first constructed a network variant (Model 2)  following the \textit{\textbf{joint} training describe-then-detect} pipeline. That is, the description network and the detection network are jointly optimized. 
Then, we developed Model 3 based on the \textit{detect-then-describe} pipeline. Specifically, the description network is combined with SIFT keypoints in Model 3.

As shown in Fig.~\ref{fig_ablation_hpatch}, with only weak supervision, the ambiguity during optimization limits the performance of \textit{\textbf{joint} training  describe-then-detect} approaches (Model 2 and DISK-W). Moreover, Model 2 is even inferior to Model 3 under viewpoint change. 
Compared to Models 2 and 3, Model 1 with our \textit{\textbf{decoupled} training describe-then-detect} pipeline produces much higher accuracy. 
This clearly demonstrates that our \textit{\textbf{decoupled} training describe-then-detect} pipeline is well suitable to weakly supervised learning to achieve superior performance.

We further test different combinations of keypoints and descriptors (Models 5-8). It can be observed that the improvement mainly comes from the descriptor, and the keypoints are slightly improved on the viewpoint change. Besides, we also illustrate the keypoints produced by our method and DISK-W in Fig.~\ref{fig_ambiguity}. DISK-W generates considerable inaccurate keypoints out of objects (\emph{e.g.}, in the sky). In contrast, our model detects more reasonable keypoints. That is because those mismatched descriptors and erroneous keypoints produced from two different components do not influence each other within our \textit{\textbf{decoupled} training describe-then-detect} pipeline.

\begin{table}[t]
    \centering
    \footnotesize
    \begin{tabular}{|c|c|c|c|}
    \hline
         \diagbox{$N_{line}$}{$w_{patch}$}  & 0.075 & 0.100 & 0.125  \\
         \hline
         75 & \makecell[c]{0.7703} & \makecell[c]{0.7705} & \makecell[c]{0.7666}\\%10.403 11.074
         \hline
         100& \makecell[c]{0.7726} & \makecell[c]{0.7748} & \makecell[c]{0.7732}\\
         \hline 
         125& 0.7732 & 0.7745 & 0.7744 \\ % ~ ~ 11.296
    \hline
    \end{tabular}
    \caption{{MMAscore achieved by our description network with different values of $N_{line}$ and $w_{patch}$ on the HPatches dataset}.}
    \label{tab_ablation_hyper}
\end{table}

\noindent\textbf{Line-to-Window Search Strategy.}
{To validate the effectiveness of our line-to-window search strategy, we developed a network variant (Model 4) by replacing our search strategy with a coarse-to-fine one (as proposed in \cite{wangLearningFeatureDescriptors2020}, illustrated in Fig. \ref{fig_l2w}(a)). For fair comparison with Model 3, SIFT keypoints are employed in this network variant. It can be observed that Model 3 outperforms Model 4 by significant margins. That is because, our line-to-window search strategy can make full use of the geometry information of camera poses to reduce the search space for accurate localization of correspondences. Consequently, higher accuracy can be achieved.}

\noindent\textbf{Number of Sampled Points $N_{line}$ and Window Size $w_{patch}$.} We conduct experiments to study the effects of $N_{line}$ and $w_{patch}$ during our line-to-window search. More sampled points and a large window size are beneficial to the performance at the expense of higher computational cost. To achieve a trade-off between performance and computational complexity, $w_{patch}=0.100$ and $N_{line}=100$ are used as the default setting.

\section{Conclusion}
In this paper, we introduce a \textit{\textbf{decoupled} training describe-then-detect} pipeline tailored for weakly supervised local feature learning. Within our pipeline, the detection network is decoupled from the description network and postponed until discriminative and robust descriptors are obtained. In addition, we propose a line-to-window search strategy to explicitly use the camera pose information to reduce search space for better descriptor learning. Extensive experiments show that our method achieves the state-of-the-art performance on three different evaluation frameworks and significantly closes the gap between fully-supervised and weakly supervised methods.

\noindent\textbf{Acknowledgement.}
This work was partially supported by the National Key Research and Development Program of China (No. 2021YFB3100800),  the Shenzhen Science and Technology Program (No. RCYX20200714114641140), and National Natural Science Foundation of China (No. U20A20185, 61972435, 62132021).

%%%%%%%%% REFERENCES
{\small
\bibliographystyle{ieee_fullname}
\bibliography{ref}
}

\end{document}

	\title{Supplementary Material for \\ Decoupling Makes Weakly Supervised Local Feature Better}
	
	\author{Kunhong~Li$^{1,2}$\quad
		Longguang~Wang$^{3}$\quad
		Li~Liu$^{3,4}$\quad 
		Qing~Ran$^5$\quad
		Kai~Xu$^3$\quad
		Yulan~Guo$^{1,2,3}$\thanks{Corresponding author: Yulan Guo (guoyulan@sysu.edu.cn).}\\
		$^1$Sun Yat-Sen University
		\quad
		$^2$The Shenzhen Campus of Sun Yat-Sen University\\
		$^3$National University of Defense Technology
		\quad
		$^4$University of Oulu
		\quad
		$^5$Alibaba Group
		% <-this % stops a space

	}
	
	% \maketitle
	% %%%%%%%%% BODY TEXT
	
	% \begin{center}
	%     \includegraphics[width=0.95\textwidth]{sup-fig/sup-model_architecture.pdf}
	%     \captionof{figure}{\label{fig-model_arch} Model architecture. The description network (a) and detection network (b) consist of several blocks, we report the scale and output channels of each block, and illustrate the details of each block in the bottom box.}
	% \end{center}
	
	\twocolumn[{%
		\maketitle
		\begin{figure}[H]
			\hsize=\textwidth % cvpr 需要
			\centering
			\includegraphics[width=0.95\textwidth]{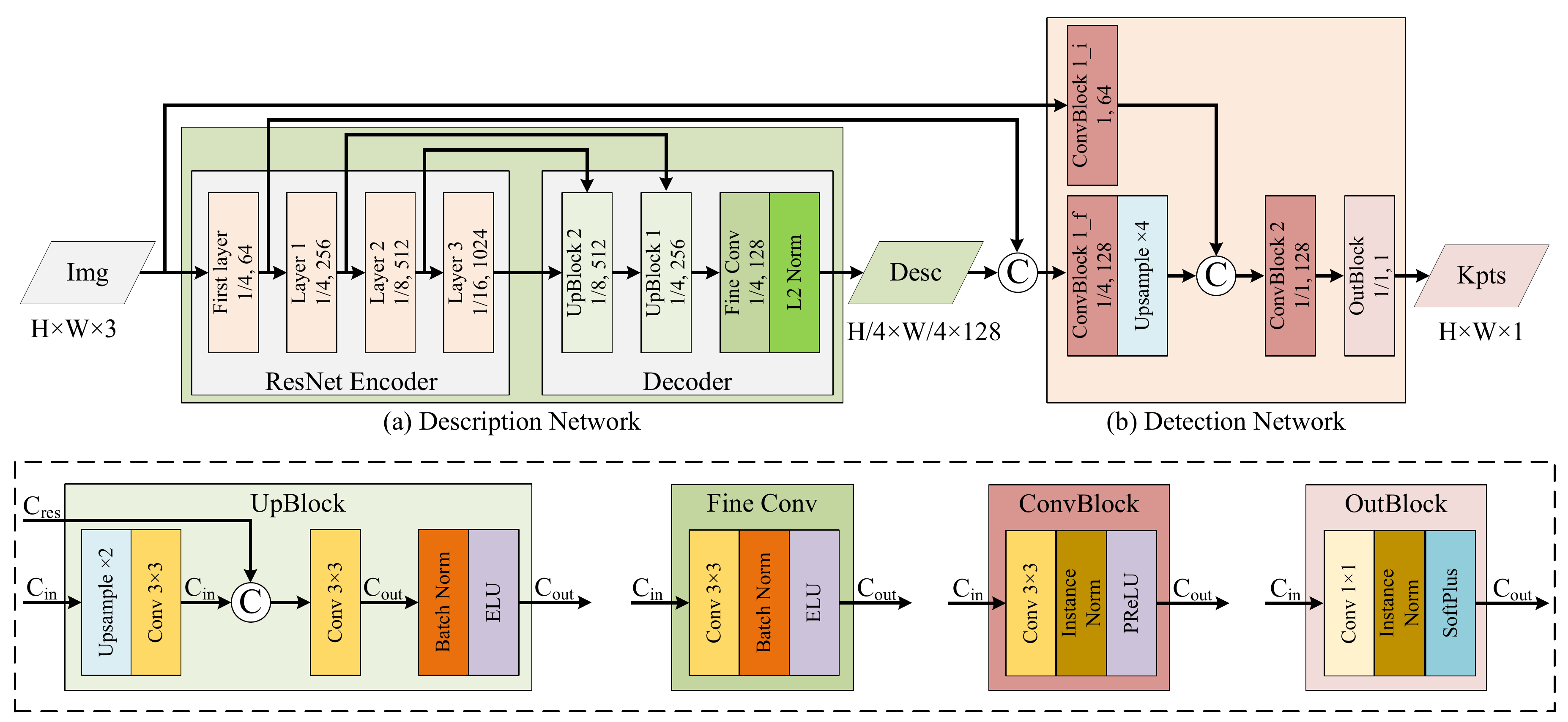}
			\caption{\label{fig-model_arch} Model architecture. The description network (a) and detection network (b) consist of several blocks, we report the scale and output channels of each block, and illustrate the details of each block in the bottom box.}
		\end{figure}
	}]

	We first introduce the details of our model in Sec.~\ref{sec-model_arch} and discuss the query points generation during description network training in Sec.~\ref{sec-random_sample}. Then, we expand the detection network training in Sec.~\ref{sec-detectnet}. Next, we show detailed experimental settings in Sec.~\ref{sec-settings}. After that, we give a discussion on the limitations and broader impact of our PoSFeat in Sec.~\ref{sec-limitation}. Finally, 
	additional qualitative results are included in Sec.~\ref{sec-vis}.

	\section{Model Architecture}
	\label{sec-model_arch}
	Our model consists of two parts, \ie the description network and the detection network, as illustrated in Fig.~\ref{fig-model_arch}. For description network, we adopts the ResUNet used in~\cite{wangLearningFeatureDescriptors2020_sup}, which follows a widely used encoder-decoder architecture. We use a truncated ResNet-50 \cite{he2016deep} (pre-trained on ImageNet \cite{deng2009imagenet}) as the encoder, and use several $3\times3$ convolution layers combining with bilinear upsampling and residual connection to construct the decoder. For detection network, we use a simple three-layer architecture. The first layer takes the original image and two feature maps from description network as inputs, and aggregate the original image and feature maps from description network for detection. For better aggregation of original image and feature maps, we use the instance normalization \cite{ulyanov2016instance} instead of batch normalization \cite{ioffe2015batch} in our detection network.
	
	\begin{figure}
		\centering
		\includegraphics[width=0.5\textwidth]{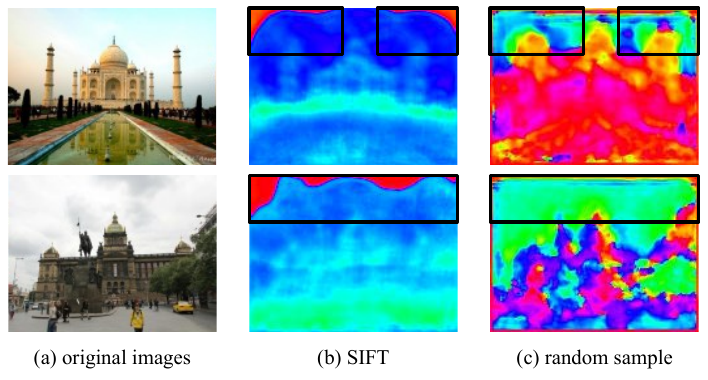}
		\caption{An illustration of pre-defined keypoints bias. We adopt PCA \cite{halko2011pca} to visualize the descriptors of the original images (a). When the description network is trained with SIFT (b), there are insufficiently trained areas (black boxes), which leads to false keypoints detection. On the contrary. When the description network is trained with grid-based random sample (c), all the areas in the image will be sufficiently trained.}
		\label{fig-random_sample}
	\end{figure}
	
	\section{Query Points Generation in Description Network Training}
	\label{sec-random_sample}
	We adopt grid-based random sampling to select query points for the training of description network to avoid the bias of pre-defined keypoints. When pre-defined keypoints (\eg SIFT) are used to train the description network, the densities of SIFT keypoints in different areas vary a lot. Consequently, areas with few SIFT keypoints are usually under optimized, as shown in Fig.~\ref{fig-random_sample}. This bias limits the discriminativeness of the descriptors and leads to detection network produces considerable false keypoints detection. To address this problem, we use grid-based random sampling to generate query points. Specifically, we first split the image into $N_g$ grids with the shape of $g\times g$. Then we uniformly select $N_g$ points with with one point in a grid. With this grid-based random sample strategy, the description network will be sufficiently trained in all areas, and thus detection network can produce more accurate keypoints.

	\section{Detection Network Training}
	\label{sec-detectnet}
	In this section, we present more details on the detection network training.
	
	As described in the main paper, we first extract the feature maps $F_1$ and $F_2$ from a image pair $I_1$ and $I_2$ with the frozen description network. Then we feed $F_1$ and $F_2$ into the detection network to produce the keypoint heatmaps, and model the keypoint distributions based on the heatmaps. Specifically, we divide these heatmaps into grids and select at most one keypoint from each grid cell. For a pixel $\boldsymbol{x}$ in image $I_1$, the probability that $\boldsymbol{x}$ is a keypoint can be formulated as,
	\begin{equation}
		P_{kp}(\boldsymbol{x}| F_1)= {\rm Softmax}(F_1^{G_{\boldsymbol{x}}})_{\boldsymbol{x}} \cdot {\rm Sigmoid}(F_1)_{\boldsymbol{x}},
	\end{equation}
	in which $F_1^{G_{\boldsymbol{x}}}$ denotes the local heatmap of the grid cell that contains pixel $\boldsymbol{x}$, ${\rm Softmax}(F_1^{G_{\boldsymbol{x}}})_{\boldsymbol{x}}$ represents the local probability of $\boldsymbol{x}$ to be a keypoint, and ${\rm Sigmoid}(F_1)_{\boldsymbol{x}}$ represents the global probability of $\boldsymbol{x}$ to be a keypoint.
	
	According to the keypoint probability distribution, we then select the candidate sets $Q_1=\{ \boldsymbol{x}_1, \boldsymbol{x}_2, \cdots | \boldsymbol{x}_i \in I_1 \}$ and $Q_2=\{ \boldsymbol{y}_1, \boldsymbol{y}_2, \cdots | \boldsymbol{y}_i \in I_2 \}$ to compute the similarity matrix $S$, whose elements are defined as,
	\begin{equation}
		S_{i,j} = F_1(\boldsymbol{x}_i)\times F_2(\boldsymbol{y}_j)^{\rm T}, \boldsymbol{x}_i \in Q_1\  \boldsymbol{y}_j \in Q_2.
	\end{equation}
	Afterwards, we can compute the matching probability $P_m$ according to the similarity matrix,
	\begin{equation}
		P_m = {\rm Softmax}(S)_1 \cdot {\rm Softmax}(S)_2,
	\end{equation}
	in which ${\rm Softmax}(S)_1$ and ${\rm Softmax}(S)_2$ denotes the softmax operation along the row and column, respectively.
	
	Next, we compute the rewards according to the epipolar constraints,
	\begin{equation}
		R( \boldsymbol{x}_i, \boldsymbol{y}_j) = 
		\begin{cases} 
			\lambda_p,  & \mbox{if }{\rm distance}( \boldsymbol{y}_j, L_{\boldsymbol{x}_i}) \leq\epsilon \\
			\lambda_n, & \mbox{if }{\rm distance}( \boldsymbol{y}_j,  L_{\boldsymbol{x}_i}) >\epsilon
		\end{cases},
	\end{equation}
	where the reward threshold $\epsilon$ is empirically set to 2. Since the description network is frozen and reliable, the matches with low matching probability are unreliable, and thus we further truncate the matching probability $P_m$ according to the reward to omit the false positive rewards for the unreliable matches. Specifically, we manually set $P_m(\boldsymbol{x}_i, \boldsymbol{y}_j)=0$ for the pairs $(\boldsymbol{x}_i, \boldsymbol{y}_j)$ whose reward $R( \boldsymbol{x}_i, \boldsymbol{y}_j)=\lambda_p$ and matching probability $P_m(\boldsymbol{x}_i, \boldsymbol{y}_j)<0.9$.
	% \begin{equation}
	%     P_m(\boldsymbol{x}_i, \boldsymbol{y}_j) = 
	%     \begin{cases}
	%     0, \mbox{if } R( \boldsymbol{x}_i, \boldsymbol{y}_j)=\lambda_p \ {\rm and}\  P_m(\boldsymbol{x}_i, \boldsymbol{y}_j) < 0.9 \\
	%     P_m(\boldsymbol{x}_i, \boldsymbol{y}_j), {\rm otherwise}
	%     \end{cases}
	% \end{equation}
	
	Finally, we compute the loss for detection network,
	\begin{equation}
		\begin{split}
			\mathcal{L}_{kp}= &-\frac{1}{|Q_1|+|Q_2|} \Big (\sum_{ \boldsymbol{x}_i, \boldsymbol{y}_j}\mathcal{L}_{rew}( \boldsymbol{x}_i, \boldsymbol{y}_j)\\
			&+ \lambda_{reg}\big(\sum_{\boldsymbol{x}_i} \log P_{kp}( \boldsymbol{x}_i) + \sum_{\boldsymbol{y}_j} \log P_{kp}( \boldsymbol{y}_j)\big) \Big ),
		\end{split}
	\end{equation}
	where $\lambda_{reg}$ is a regularization penalty and the reward loss $\mathcal{L}_{rew}( \boldsymbol{x}_i, \boldsymbol{y}_j)$ is defined as:
	\begin{equation}
		\mathcal{L}_{rew}( \boldsymbol{x}_i, \boldsymbol{y}_j) \!=\! P_{m}( \boldsymbol{x}_i, \boldsymbol{y}_j)\cdot R( \boldsymbol{x}_i, \boldsymbol{y}_j) \cdot \log (P_{kp}( \boldsymbol{x}_i)P_{kp}( \boldsymbol{y}_i)).
		%         &\cdot (\log P( {x}_i) + \log P( {y}_j))
	\end{equation}
	Note that, the $P_m$ is truncated according to the rewards and thus the match pairs with positive rewards but low matching probability are left neutral. 
	
	\begin{figure*}
		\centering
		\includegraphics[width=0.95\textwidth]{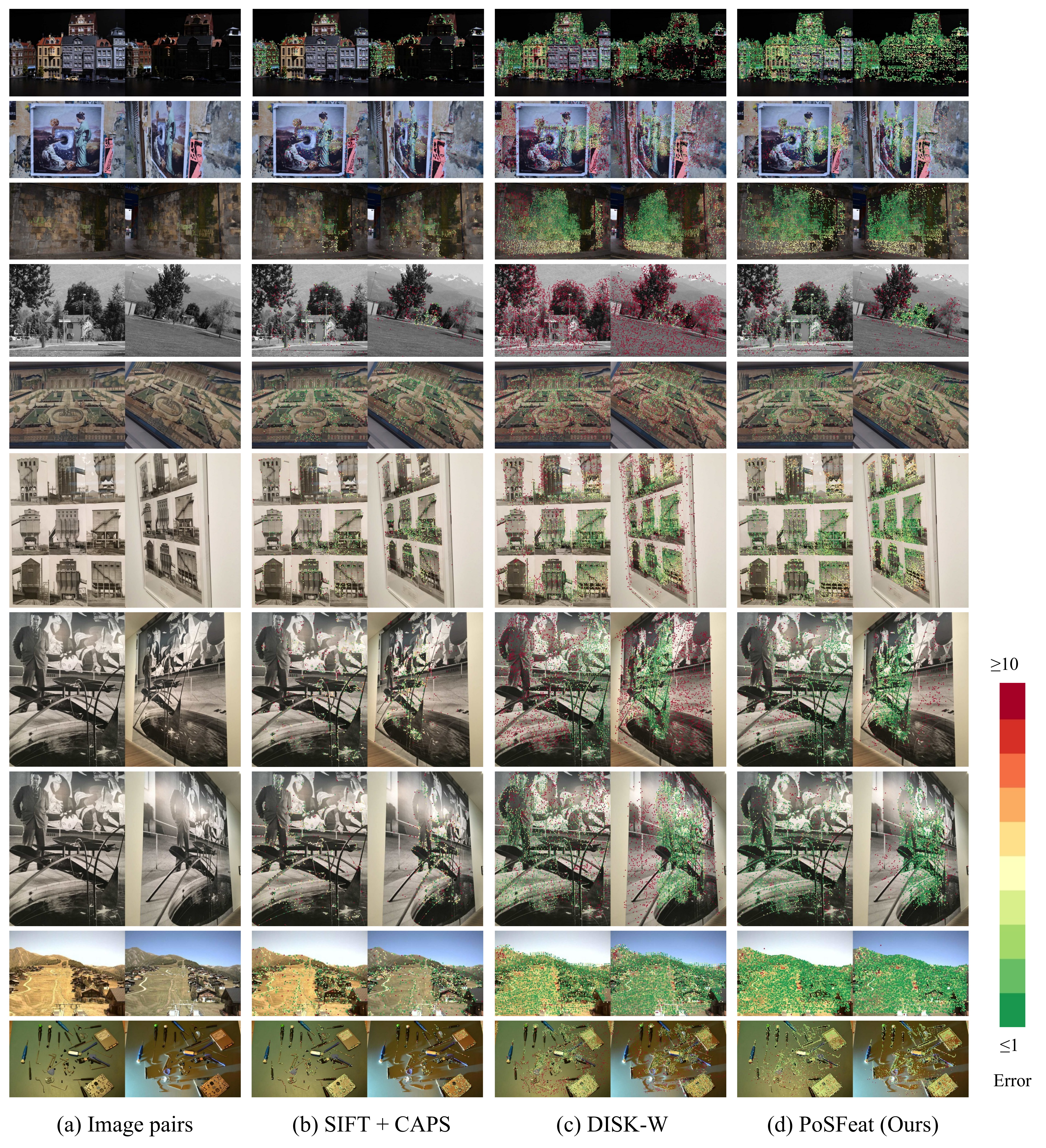}
		\caption{Qualitative results on HPatches. The same as figures in main paper, only successfully matched keypoints are shown and colored according to their match errors. The colorbar is shown on the right. Best viewed in color.}
		\label{fig-hpatches_more}
	\end{figure*}
	
	\section{Experimental Settings}
	\label{sec-settings}
	In this section, we present the hyper-parameters of our method on different datasets. During inference, we apply non-maximum suppression (NMS) to detect keypoints, and use a mutual nearest neighbour matcher for matching. Instead of resizing the images, we crop the images from the top-left side to guarantee both the height and width of the images are divisible by 16. %Therefore, the extraction is applied with the original resolution images.
	
	\paragraph{\textbf{HPatches Dataset}~\cite{hpatches_2017_cvpr_sup}.} The NMS size is set to be $3\times 3$ due to the existence of low-resolution images, and the maximum keypoint numbers are limited to be 8192. 
	
	\paragraph{\textbf{Aachen Day-Night Dataset}~\cite{zhang2021reference_sup}.} Because of the high image resolutions, the NMS size is set to be $7\times 7$ on the Aachen Day-Night dataset, and the maximum keypoint numbers are limited to be $16k$. Note that, keypoints with scores smaller than 0.9 in the heatmaps are filtered out.
	
	\paragraph{\textbf{ETH Local Feature Benchmark}~\cite{schonberger2017comparative_sup}.} The NMS size is set to be $7\times 7$, and the maximum keypoints numbers are limited to be $20k$. Keypoints with scores smaller than 0.9 in the heatmaps are also filtered out. We additionally applying ratio test during matching with a threshold 0.8 to achieve robust reconstruction. 
	
	\section{Limitations and Broader Impact}
	\label{sec-limitation}
	The PoSFeat suffers limited capability to deal with large rotation and scale changes. On the HPatches dataset, our PoSFeat produces limited performance on several scenes with pure rotation. On the ETH local feature benchmark, our PoSFeat cannot well handle the scenes with extreme scale changes thus has limited performance in \#Imgs (\eg, only 419 images are registered in Mardrid Metropolis).
	
	The PoSFeat is a general local feature method, although we only apply it with image matching, visual localization and 3D reconstruction in our paper, it can be easily extended to recognize or reconstruct human faces. Therefore, the researches and the applications about the recognition or reconstruction of human faces must strictly respect the personality rights and privacy regulations.
	
	\begin{figure}
		\centering
		\includegraphics[width=0.5\textwidth]{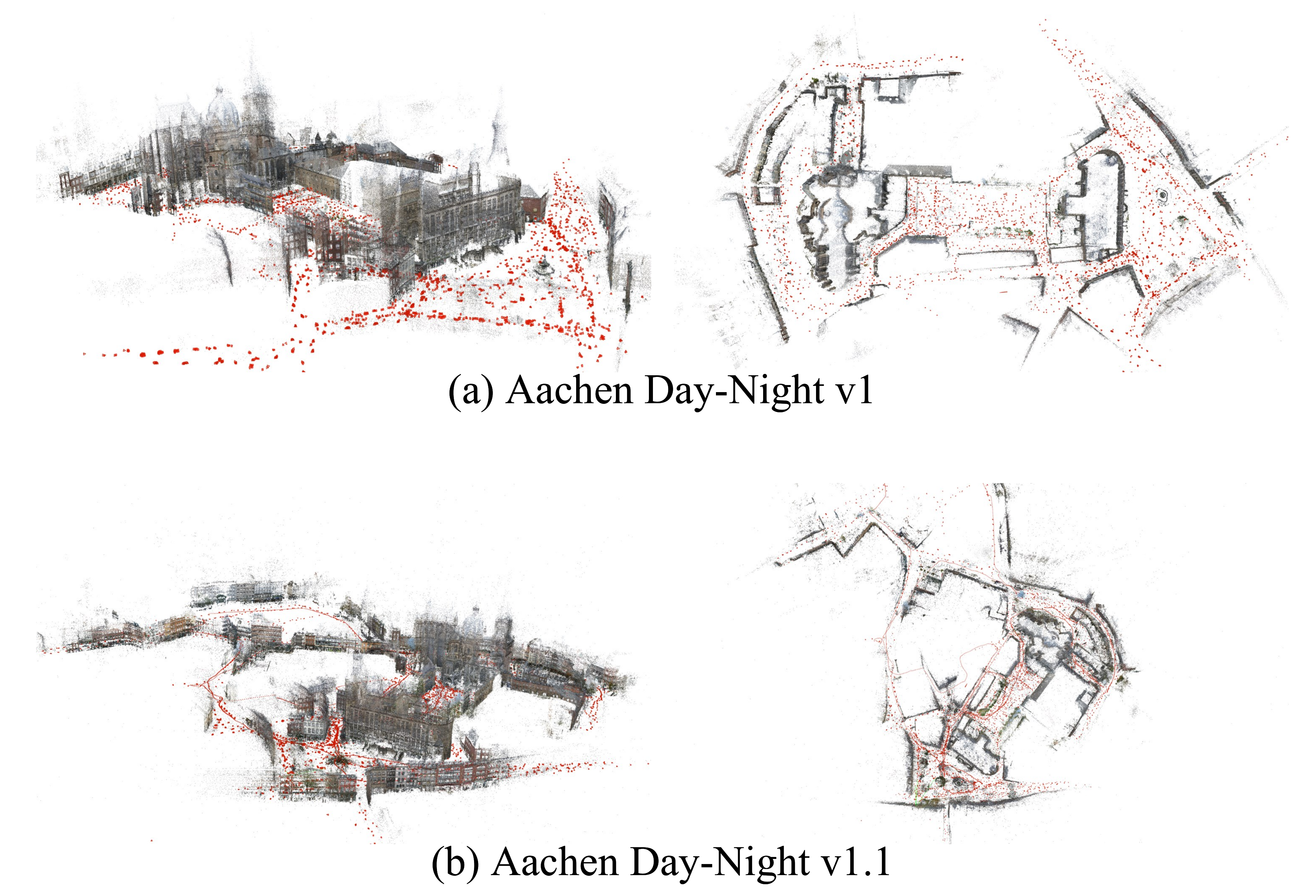}
		\caption{The sparse 3D models of Aachen. These models are reconstructed using Colmap~\cite{schonberger2016structure_sup} with features extracted by PoSFeat, and are further used to do night-time images localization. Note that these models are reconstructed based on the camera poses provided by the author of the dataset.}
		\label{fig-aachen_rec}
	\end{figure}
	
	\begin{figure}
		\centering
		\includegraphics[width=0.5\textwidth]{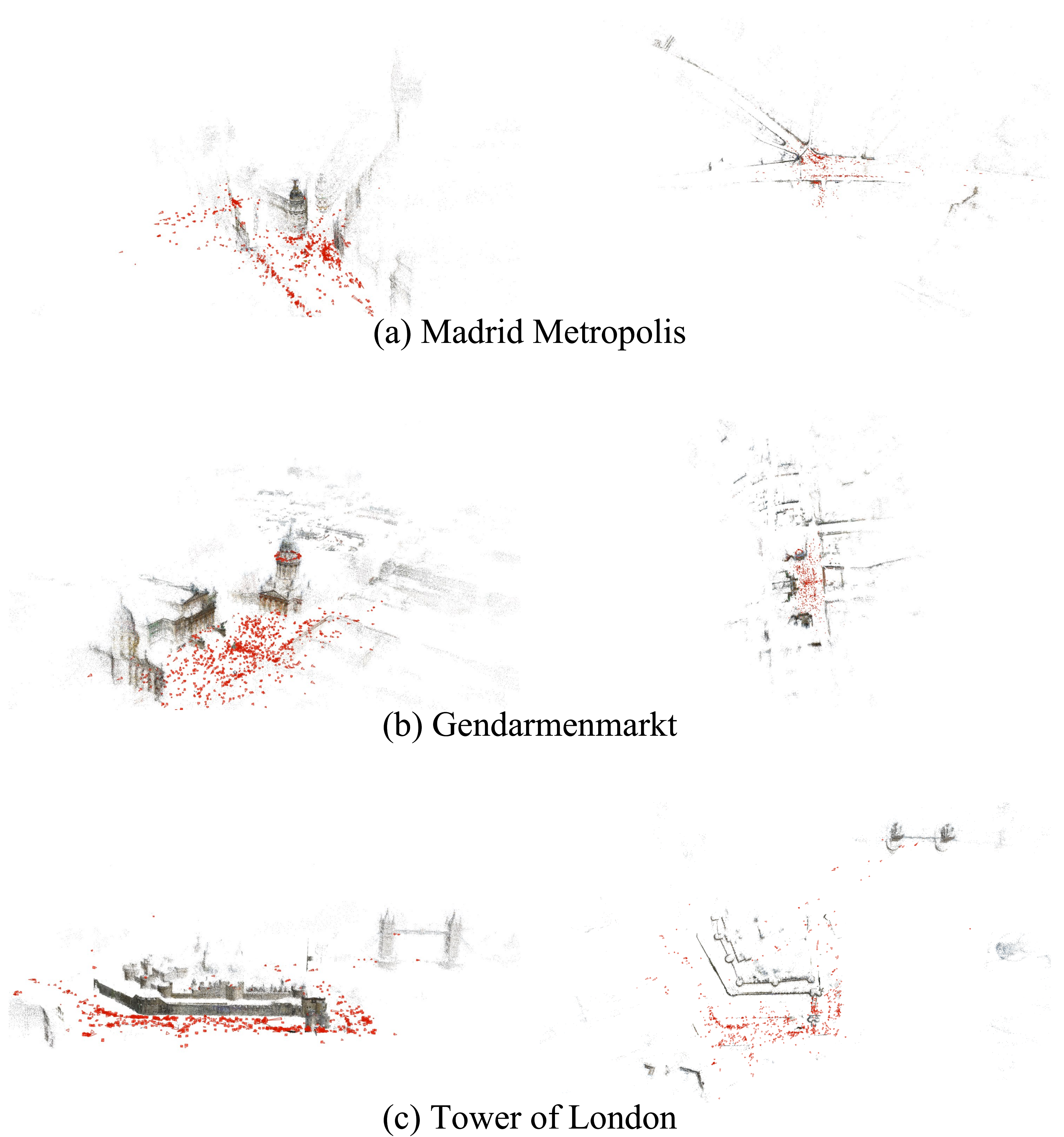}
		\caption{The sparse 3D reconstruction results on ETH local feature benchmark.}
		\label{fig-ETH}
	\end{figure}
	
	\section{Visualization}
	\label{sec-vis}
	We present some qualitative results in this section. The image matching results on HPatches are shown in Fig.~\ref{fig-hpatches_more}. The 3D models of Aachen are illustrated in Fig.~\ref{fig-aachen_rec}, which is reconstructed with the features extracted by PoSFeat, and is used to do visual localization on Aachen Day-Night dataset . And the 3D reconstruction results on ETH local feature benchmark are shown in Fig.~\ref{fig-ETH}.

	%%%%%%%%% REFERENCES
	{\small
		\bibliographystyle{ieee_fullname}
		\bibliography{arxiv_sup}
	}